\pdfoutput=1

\documentclass[11pt]{article}
\RequirePackage[OT1]{fontenc}
\RequirePackage{amsthm,amsmath}
\RequirePackage[numbers]{natbib}
\RequirePackage[colorlinks,citecolor=blue,urlcolor=blue]{hyperref}

\usepackage{indentfirst}
\usepackage[utf8]{inputenc} 
\usepackage[T1]{fontenc}    
\usepackage[colorlinks]{hyperref} 
\usepackage{url}            
\usepackage{booktabs}       
\usepackage{amsfonts}       
\usepackage{amsmath}
\usepackage{nicefrac}       
\usepackage{microtype}      
\usepackage{color}
\usepackage{bbm}
\usepackage{amssymb, enumerate}
\usepackage{makecell}
\usepackage{wrapfig}
\usepackage{extarrows}
\usepackage{arydshln}

\usepackage{caption}
\usepackage{multirow}

\usepackage{amsmath,amsfonts,amsthm, amssymb, enumerate}
\usepackage{lipsum}
\usepackage{caption}
\usepackage{amssymb,graphicx,subfigure}
\usepackage[bottom]{footmisc}
\usepackage[dvipsnames]{xcolor}
\usepackage{array,multirow}
\usepackage{enumitem}

\usepackage{microtype}
\usepackage{graphicx}
\usepackage{subfigure}
\usepackage{booktabs} 
  
\usepackage{amsmath}
\usepackage{amssymb}
\usepackage{booktabs}  
\usepackage{times}
\usepackage{epsfig} 
\usepackage{amsfonts, amsthm, enumerate} 
\usepackage[bottom]{footmisc} 
\usepackage{enumitem}
\usepackage{setspace}
\usepackage{wrapfig}
\usepackage{extarrows}
\usepackage{makecell}
\usepackage{bbm}  
\usepackage{textcomp}
\usepackage{xcolor} 
\usepackage{caption} 

\usepackage{booktabs} 
\usepackage{siunitx}
\usepackage{multirow} 
\usepackage{caption} 
\usepackage[switch]{lineno} 
\usepackage{amsmath}
\usepackage{amssymb}
\usepackage{graphicx}

\usepackage[font=small,labelfont=bf]{caption}

\usepackage{algorithm,algpseudocode}
\algnewcommand\TR{\item[{\textbf{Training phase}}]}
\algnewcommand\TE{\item[{\textbf{Test phase}}]}
\algnewcommand\Input{\item[{{Input:}}]}
\algnewcommand\Output{\item[{{Output:}}]}
\algnewcommand\Initialize{\item[{{Initialize:}}]}
\algnewcommand{\return}[1]{
	\State \textbf{return:}
	\Statex \hspace*{\algorithmicindent}\parbox[t]{.8\linewidth}{\raggedright #1}
}


\usepackage{setspace}
\begin{document}
	\title{Integrating ChatGPT into Secure Hospital Networks: A Case Study on Improving Radiology Report Analysis}
	\date{}
	\author{
		Kyungsu Kim$^{1}$\thanks{Equal contribution}, Junhyun Park$^{2}$\footnotemark[1], Saul Langarica$^{1}$, Adham Mahmoud Alkhadrawi$^{1}$, Synho Do$^{1}$\thanks{Correspondence to}\\ 
		\\
		{\footnotesize $^{1}$Massachusetts General Hospital and Harvard Medical School, Boston, MA, USA, 02115}\\
		{\footnotesize $^{2}$Department of Robotics and Mechatronics Engineering, DGIST, Daegu, Republic of Korea, 42988} 
	} 
 
	\maketitle

\begin{abstract} 
This study demonstrates the first in-hospital adaptation of a cloud-based AI, similar to ChatGPT, into a secure model for analyzing radiology reports, prioritizing patient data privacy. By employing a unique sentence-level knowledge distillation method through contrastive learning, we achieve over 95\% accuracy in detecting anomalies. The model also accurately flags uncertainties in its predictions, enhancing its reliability and interpretability for physicians with certainty indicators. These advancements represent significant progress in developing secure, efficient AI tools for healthcare, suggesting a promising future for in-hospital AI applications with minimal supervision.
\end{abstract}

\section{Introduction}
\label{introduction}
The research explores the integration of artificial intelligence (AI), specifically large language models (LLMs) like ChatGPT into radiology within hospitals with an emphasis on maintaining security during implementation.

Despite the proven effectiveness of these AI tools in processing radiological reports \citep{wu2024benchmarking,mirza2024using,lee2023benefits}, their integration into hospital environments poses challenges due to the sensitive nature of patient data and the need for data confidentiality \citep{senbekov2020}. The direct use of cloud-based LLMs like ChatGPT is limited by data security concerns, especially when considering healthcare regulations such as HIPAA \citep{gostin2009beyond} and GDPR \citep{voigt2017eu}. 

Our study addresses this by adapting these LLMs for secure, internal use within hospital radiology departments, transforming them into closed-network systems to comply with healthcare privacy standards. This approach aims to leverage the advanced capabilities of LLMs while safeguarding patient data privacy. 

This paper delves into how radiology reports can be automatically classified as normal or abnormal using cloud-based/high-performing LLMs like ChatGPT, with the goal of adapting these models for secure and internal use within hospital networks. This approach aims to enhance hospital workflows by streamlining the analysis of radiology findings, potentially leading to more efficient and accurate medical diagnostics and patient care management.

This investigation is important for enhancing the practical utility of AI in radiology, ensuring both technological advancement and adherence to the paramount principle of patient confidentiality. Our contribution is three-fold:
\begin{itemize}
    \item Successfully adapted a cloud-based model like ChatGPT into an on-site version with over 95\% accuracy for detecting anomalies in radiology reports, offering a secure method for local data processing.
    \item Demonstrated that sentence-level knowledge distillation outperforms traditional document-level methods in improving model replication by better identifying rare abnormal findings, supported by analytical evidence.
    \item Improved model interpretability by adding an ``uncertain" label to the usual ``normal" and ``abnormal" in sentence-level knowledge distillation. This allows the model to identify ambiguous cases in radiology reports, enhancing sentence-level accuracy and clarity. The provided code visualizes sentence-based predictions, helping physicians focus on critical findings during review by clearly marking uncertain sentences.
\end{itemize}

\begin{table*}[t]
\centering
  {\resizebox{1.0\linewidth}{!}{%
  {\begin{tabular}{ccccccc} 
    \toprule
    Study                & \begin{tabular}[c]{@{}c@{}}Does it demonstrate the \\feasibility of knowledge \\distillation (KD) learning \\from a cloud-based model \\to an on-premises model?\end{tabular} & \begin{tabular}[c]{@{}c@{}}Does it propose an \\advancement technique \\for KD and provide its\\~reasons of improvement?\end{tabular} & \begin{tabular}[c]{@{}c@{}}Model type\\(Cloud \\or On-premises type)?\end{tabular}                                    & \begin{tabular}[c]{@{}c@{}}Does it address model \\learning? / If yes, what \\kind of learning data\\~does the model use \\among public data (p), \\private data (i), cloud model’s\\~prediction result data (c)?\end{tabular} & \begin{tabular}[c]{@{}c@{}}What is the showcase \\application of the model?\end{tabular}        & \begin{tabular}[c]{@{}c@{}}Is the public code \\ available?\end{tabular}  \\ 
    \midrule
    \cite{li2023decoding}        & No                                                                                                                                                                            & No                                                                                                                                    & Cloud (without KD)                                                                                                                & No                                                                                                                                                                                                                             & Summarization                                                                                   & No                                                                       \\
    \cite{liu2023evaluating}       & No                                                                                                                                                                            & No                                                                                                                                    & Cloud or On-premises (without KD)                                                                                                 & No                                                                                                                                                                                                                             & Generation                                                                                      & No                                                                       \\
    \cite{ma2023impressiongpt}       & No                                                                                                                                                                            & No                                                                                                                                    & Cloud  (without KD)                                                                                                               & No                                                                                                                                                                                                                             & Generation                                                                                      & Yes                                                                      \\
    \cite{Liu2023RadiologyLlama2BL}      & No                                                                                                                                                                            & No                                                                                                                                    & On-premises (without KD)                                                                                                          & Yes (p)                                                                                                                                                                                                                        & Generation                                                                                      & No                                                                       \\
    \cite{liu2023radiology}       & No                                                                                                                                                                            & No                                                                                                                                    & On-premises   (without KD)                                                                                                        & Yes (p)                                                                                                                                                                                                                        & Generation                                                                                      & (Param. only)                                                            \\
    \cite{zhong2023chatradio}     & No                                                                                                                                                                            & No                                                                                                                                    & On-premises    (without KD)                                                                                                       & Yes (i)                                                                                                                                                                                                                        & Generation                                                                                      & No                                                                       \\
    \cite{van2023radadapt}       & No                                                                                                                                                                            & No                                                                                                                                    & On-premises   (without KD)                                                                                                        & Yes (p)                                                                                                                                                                                                                        & Generation                                                                                      & Yes                                                                      \\
    \cite{mukherjee2023feasibility}  & No                                                                                                                                                                            & No                                                                                                                                    & On-premises  (without KD)                                                                                                         & Yes (p)                                                                                                                                                                                                                        & Classification                                                                                  & No                                                                       \\
    \cite{bressem2020highly}    & No                                                                                                                                                                            & No                                                                                                                                    & On-premises  (without KD)                                                                                                         & Yes (i)                                                                                                                                                                                                                        & Classification                                                                                  & Yes                                                                      \\
    \cite{yan2022radbert}     & No                                                                                                                                                                            & No                                                                                                                                    & On-premises  (without KD)                                                                                                         & Yes (p)                                                                                                                                                                                                                        & Classification                                                                                  & (Param. only)                                                            \\ 
    \midrule
    \textbf{~ Our study} & \textbf{Yes}                                                                                                                                                                  & \begin{tabular}[c]{@{}c@{}}\textbf{Yes }\\\textbf{(i.e., Sentence-level KD)}\end{tabular}
    & \begin{tabular}[c]{@{}c@{}}\textbf{On-premises} \\\textbf{(trained with KD }\\\textbf{from cloud model)}\end{tabular} & \textbf{Yes (c)}                                                                                                                                                                                                               & \begin{tabular}[c]{@{}c@{}}\textbf{Classification} \\\textbf{(Abnormal detection)}\end{tabular} & \textbf{Yes}                                                             \\
    \toprule
    \end{tabular}
    }}}
      \caption{Comparison of our study with related ones to develop language models applied to electronic medical record (EMR) documents. Our is the first study aimed at reproducing the cloud model into a non-cloud/secure model (second column).}
      \label{tab:related_works}
\end{table*}

\section{Related Works}
\label{sec:rel}
LLMs, like ChatGPT, are used in radiology research to analyze reports and are classified as cloud-based or non-cloud-based. Rows 1-3 of Table \ref{tab:related_works} illustrate how cloud-based studies employ LLMs for report generation or summarization, utilizing prompt engineering without additional model training. However, this dependence on cloud storage raises concerns regarding data security. On the other hand, as shown in column 5 of Table \ref{tab:related_works}, non-cloud (i.e., on-premises) approaches, which are described in rows 4–10 of the table, need human annotation for model training, which means a substantial amount of work for data preparation—more particularly, \textit{p}- or \textit{i}-type annotation.

\begin{figure}[t] 
\centering
  {\includegraphics[width=0.8\linewidth]{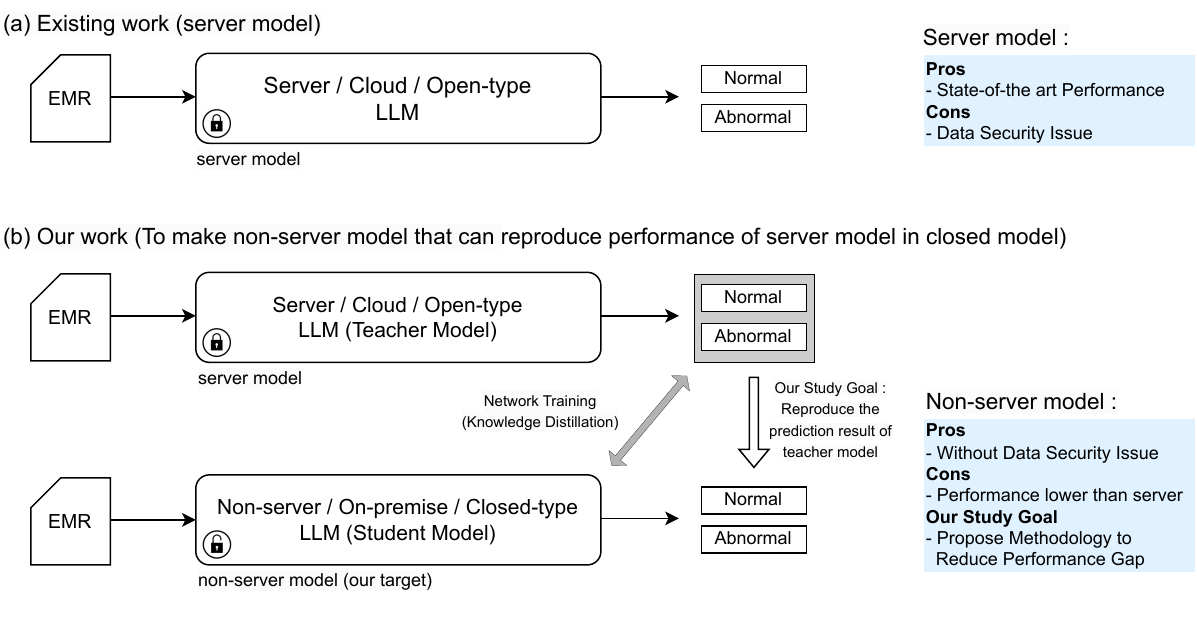}}  
  \caption{Our study is the first to test the feasibility of distilling knowledge from cloud model like ChatGPT into a non-cloud model for radiology report analysis}
  \label{fig:kd}
\end{figure}

In contrast to these previous studies that focus on either cloud type or non-cloud type, our study is the first to use both types, by incorporating knowledge distillation (KD) \citep{gou2021knowledge} into LLMs for radiology report analysis (see the second column in Table \ref{tab:related_works} and Fig. \ref{fig:kd}). Specifically, we utilize both types by replicating the cloud type as the non-cloud type through KD. This approach involves training a condensed non-cloud model (referred to as the student) to emulate the capabilities of a more extensive cloud model (known as the teacher), such as ChatGPT.  Our approach stands out for utilizing automatically processed data from the cloud model to train the non-cloud model (i.e., as \textit{c}-type data), instead of relying on human-annotated data such as \textit{p}- or \textit{i}-type annotation data stated in rows 4-10 in column 5 of Table \ref{tab:related_works}. Accordingly, our approach bypasses the requirement for data labeled by humans.  Furthermore, our technique addresses the security concerns associated with evaluation data, by uploading only the limited training data to the cloud model, thereby excluding the remaining/unlimited evaluation data that instead utilizes our trained non-cloud model without the security concerns.

\begin{figure}[t] 
\centering
{\includegraphics[width=0.8\linewidth]{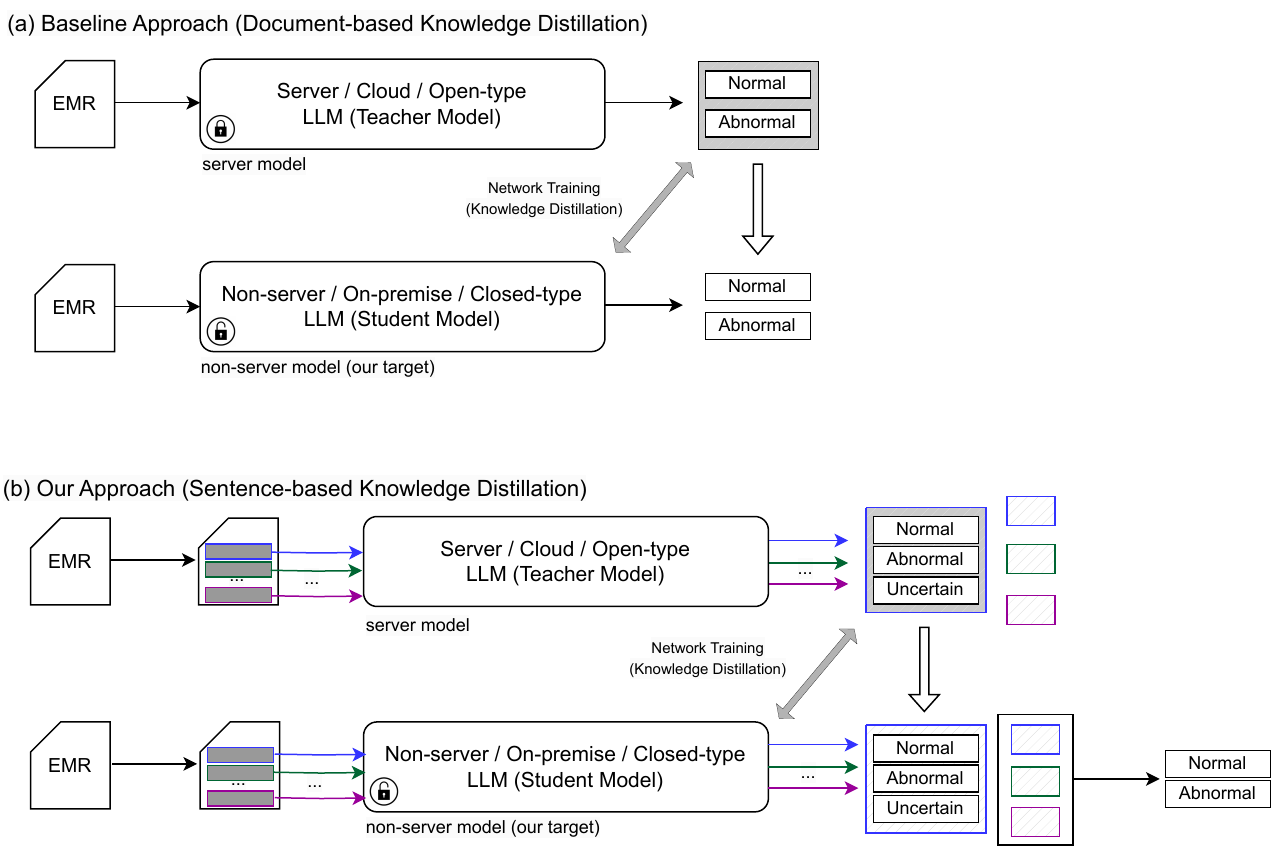}}
  \caption{Improving knowledge distillation performance and interpretability, our approach incorporates sentence-level knowledge distillation and enhances reliability by introducing an additional label (uncertain) for the network explicitly to indicate uncertainty in prediction results}
  \label{fig:dvss}
\end{figure}

In addition, we have developed an advanced technique for KD using a sentence-level approach. This method outperforms standard KD methods as shown in the third column of Table \ref{tab:related_works} and Fig. \ref{fig:dvss}. This new approach greatly enhances the model's ability to identify anomalies in documents, especially in challenging scenarios when the document includes a lower number of abnormal sentences. Furthermore, the incorporation of contrastive learning loss into the KD process has enhanced the model's precision in recognizing the class with few training instances (i.e., normal class). Therefore, our incorporation of sentence-level KD and contrastive learning loss results in a notable improvement in the utilization of KD for analyzing radiological reports in language models.

\section{Method}
\label{sec:method}
In this section, we introduce a KD approach for anomaly detection in radiology reports. This involves two primary methods: the baseline document-level KD (Sec. \ref{sec:based_doc_kd}) and our proposed sentence-level KD (Sec. \ref{sec:based_sent_kd}) approaches.

\subsection{Baseline: Document-level Knowledge Distillation}\label{sec:based_doc_kd}
In this section, we present a method for anomaly detection (AD) in radiology reports using document-based knowledge distillation (D-KD), a widely used KD technique. As there are no cases in which the KD technique has been applied to processing radiology reports, we proposed this baseline technique as a representative and straightforward method for this purpose. Sec. \ref{sec:based_doc_gtrd} details the process of extracting labels from the teacher model, ChatGPT, to create training data.  
Sec. \ref{sec:based_doc_kd_tr} details the KD training of an on-premise LLM student model using data from Sec. \ref{sec:based_doc_gtrd}, showing the subsequent testing of the model.

\subsubsection{Label Extraction from Teacher Model}\label{sec:based_doc_gtrd}
The radiology report was input into the teacher model (i.e., ChatGPT), which then determined if the report was normal or abnormal, using $n$ for normal labels and $a$ for abnormal. This process is defined by the function $f^d$ as follows:
\begin{align}\label{eq:f_d}
f^d(x_i;c) =: y^d_i \in \{a,n\}.
\end{align}
Here, $x_i$ represents the $i$-th radiology document, and $c$ denotes the question prompt (see the details in Appendix) used in ChatGPT to generate predictions for anomaly detection by the binary label symbolized as $y^d_i$. Through this process, $T$ number of KD-training data pairs for AD (i.e., $D = (x_i,y^d_i)_{i=1}^{T}$) were constructed.

\subsubsection{Training and Inference for Student Model}\label{sec:based_doc_kd_tr}

\begin{itemize}
\item \textbf{Training Phase.} We updated the student model $g_{\theta}$ from the training data constructed in Sec. \ref{sec:based_doc_gtrd} with model parameters $\theta$ to minimize the objective below
\begin{align}\label{loss_doc} 
 \theta^* \leftarrow \min_{\theta} \, \mathbb{E}_{(i \in \{1:T\})} \bigg[\mathcal{L}_{\theta}\Big(g_{\theta}(x_i),y^d_i\Big)\bigg]
\end{align} 
where $\theta^*$ is the trained parameter and $\mathcal{L}_{\theta}$ is our objective function for KD (see the details in Sec. \ref{sec:kd_loss}). 

\item \textbf{Inference Phase. }  Binary classification evaluation of radiology reports for AD was performed from the student model $g_{\theta^*}$ on which KD-learning was completed as follows:
\begin{align}\nonumber 
 p_{a} &\leftarrow  \, g_{\theta^*}(x^{te})_{\{a\}}, \\\nonumber 
p_{n} &\leftarrow  1 - p_{a}.
\end{align}
Here, $x^{te}$ denotes the radiology report used for testing, and the student model $g_{\theta^*}(x^{te})$ converts it into a binary probability vector $(p_{a},p_{n})$ within $\mathbb{R}^2$ (i.e.,  $g_{\theta^*}(x^{te}) \mapsto (g_{\theta^*}(x^{te})_{\{a\}},g_{\theta^*}(x^{te})_{\{n\}})=:(p_{a},p_{n}) \in [0,1]^{2}$). This vector's first element, $p_{a}$, reflects the model's estimated probability of the input document $x^{te}$ being abnormal.
\end{itemize}

\subsection{Proposed: Sentence-level Knowledge Distillation}\label{sec:based_sent_kd}
In this section, we newly introduce S-KD, a sentence-level-based KD method, more advanced than D-KD in Sec. \ref{sec:based_doc_kd}.
\subsubsection{Label Extraction from Teacher Model}
Unlike the baseline method, where the entire radiology report is input into the teacher model as shown in Eq. \eqref{eq:f_d}, our approach inputs individual sentences $s_{ij} \in \{s_{ij}\}_{j=1}^{D}$ from report $x_i$ into ChatGPT. This yields ternary anomaly detection (AD) labels $\{a,n,u\}$, explicitly incorporating an ``uncertain'' label $u$ alongside the existing binary labels $\{a,n\}$:
\begin{align}\nonumber 
f^s(s_{ij};c) =: y^s_{ij} \in \{a,n,u\}.
\end{align}
Here, $x_i$, $s_{ij}$, and $y^s_{ij}$ denote the $i$-th radiology document, its $j$-th sentence, and its model prediction as ternary label, respectively.
Accordingly, a total $\sum_{j=1}^T D_j$ of KD-training data pairs for AD were constructed, as $({s_{ij},y^s_{ij}})_{(i,j)=(1,1)}^{(T,D_i)}$.

\begin{table}[h]
\centering  
  {\resizebox{0.6\linewidth}{!}{%
  {\begin{tabular}{ccccc} 
\toprule
\textbf{Model}                                                           & \textbf{Accuracy}                                                & \textbf{\textbf{Specificity}}                                     & \textbf{Sensitivity}                                              & \textbf{\textbf{AUC}}                                               \\ 
\toprule
\begin{tabular}[c]{@{}c@{}}RadBERT-Roberta\\-4m-document\end{tabular}    & 85.52                                                            & 0.858                                                             & 0.84                                                              & 0.901                                                               \\
\begin{tabular}[c]{@{}c@{}}RadBERT-Roberta\\-4m-sentence\end{tabular}    & \begin{tabular}[c]{@{}c@{}}\textbf{95.06}\\(+ 9.54)\end{tabular} & \begin{tabular}[c]{@{}c@{}}\textbf{0.941}\\(+ 0.083)\end{tabular} & \begin{tabular}[c]{@{}c@{}}\textbf{0.952}\\(+ 0.112)\end{tabular} & \begin{tabular}[c]{@{}c@{}}\textbf{0.977} \\(+ 0.076)\end{tabular}  \\ 
\midrule
\begin{tabular}[c]{@{}c@{}}BioMed-Roberta\\-document\end{tabular}        & 86.12                                                            & 0.82                                                              & 0.869                                                             & 0.877                                                               \\
\begin{tabular}[c]{@{}c@{}}BioMed-Roberta\\-sentence\end{tabular}        & \begin{tabular}[c]{@{}c@{}}\textbf{94.6}\\(+ 8.48)\end{tabular}  & \begin{tabular}[c]{@{}c@{}}\textbf{0.947}\\(+ 0.127)\end{tabular} & \begin{tabular}[c]{@{}c@{}}\textbf{0.943}\\(+ 0.074)\end{tabular} & \begin{tabular}[c]{@{}c@{}}\textbf{0.979}\\(+ 0.102)\end{tabular}   \\ 
\midrule
\begin{tabular}[c]{@{}c@{}}BlueBERT\\-document\end{tabular}              & 91.17                                                            & 0.91                                                              & 0.922                                                             & 0.958                                                               \\
\begin{tabular}[c]{@{}c@{}}BlueBERT\\-sentence\end{tabular}              & \begin{tabular}[c]{@{}c@{}}\textbf{93.43}\\(+ 2.26)\end{tabular} & \begin{tabular}[c]{@{}c@{}}\textbf{0.933}\\(+ 0.023)\end{tabular} & \begin{tabular}[c]{@{}c@{}}\textbf{0.945}\\(+ 0.023)\end{tabular} & \begin{tabular}[c]{@{}c@{}}\textbf{0.98}\\(+ 0.022)\end{tabular}    \\ 
\midrule
\begin{tabular}[c]{@{}c@{}}Clinical BERT\\-document\end{tabular}         & 90.15                                                            & 0.888                                                             & 0.961                                                             & 0.968                                                               \\
\begin{tabular}[c]{@{}c@{}}Clinical BERT\\-sentence\end{tabular}         & \begin{tabular}[c]{@{}c@{}}\textbf{93.07}\\(+ 2.92)\end{tabular} & \begin{tabular}[c]{@{}c@{}}\textbf{0.922}\\(+ 0.034)\end{tabular} & \begin{tabular}[c]{@{}c@{}}\textbf{0.973}\\(+ 0.012)\end{tabular} & \begin{tabular}[c]{@{}c@{}}\textbf{0.982}\\(+ 0.014)\end{tabular}   \\ 
\midrule
\begin{tabular}[c]{@{}c@{}}BiomedBERT\\-document\end{tabular}            & 92.76                                                            & \textbf{0.93}                                                     & 0.916                                                             & 0.926                                                               \\
\begin{tabular}[c]{@{}c@{}}BiomedBERT\\-sentence\end{tabular}            & \begin{tabular}[c]{@{}c@{}}\textbf{93.07}\\(+ 0.31)\end{tabular} & \begin{tabular}[c]{@{}c@{}}0.926\\(- 0.004)\end{tabular}          & \begin{tabular}[c]{@{}c@{}}\textbf{0.961}\\(+ 0.045)\end{tabular} & \begin{tabular}[c]{@{}c@{}}\textbf{0.982}\\(+ 0.056)\end{tabular}   \\ 
\midrule
\begin{tabular}[c]{@{}c@{}}BioBERT\\-document\end{tabular}               & 90.5                                                             & 0.905                                                             & 0.906                                                             & 0.959                                                               \\
\begin{tabular}[c]{@{}c@{}}BioBERT\\-sentence\end{tabular}               & \begin{tabular}[c]{@{}c@{}}\textbf{92.37}\\(+ 1.87)\end{tabular} & \begin{tabular}[c]{@{}c@{}}\textbf{0.923}\\(+ 0.018)\end{tabular} & \begin{tabular}[c]{@{}c@{}}\textbf{0.929}\\(+ 0.023)\end{tabular} & \begin{tabular}[c]{@{}c@{}}\textbf{0.973}\\(+ 0.014)\end{tabular}   \\ 
\midrule
p-value                                                                  & 0.002                                                            & 0.04                                                              & 0.002                                                             & 0.002                                                               \\ 
\midrule
\begin{tabular}[c]{@{}c@{}}\textcolor{black}{~Average ratio of}\\\textcolor{black}{sent./doc. performance}\end{tabular} & \textcolor{black}{1.047}                                                            & \textcolor{black}{1.053}                                                             & \textcolor{black}{1.053}                                                             & \textcolor{black}{1.051}                                                               \\
\bottomrule
\end{tabular}
}}}
  \caption{Anomaly detection performance comparison between document-level and sentence-level (ours) KD approaches across various backbone student models}
  \label{tab:comp_d_s}
\end{table}

\subsubsection{Training and Inference for Student Model}  
\begin{itemize}
\item \textbf{Training Phase. } Our KD training follows the same method as document-level KD training in Eq. \eqref{loss_doc}, but differs only in the input/label data as sentence-level:
\begin{align}\label{loss_sent} 
 \theta^* \leftarrow \min_{\theta} \, \mathbb{E}_{(i \in T, \, j \in D_i)} \bigg[\mathcal{L}_{\theta}\Big(g_{\theta}(s_{ij}), y^s_{ij}\Big)\bigg].
\end{align}
\item \textbf{Inference Phase. } Therefore, the learned student model $g_{\theta^*}$ can provide ternary classification prediction results for individual sentences in the test report $x^{te}$, i.e., $g_{\theta^*}(s^{te}_j) \mapsto (p^j_{a},p^j_{n},p^j_{u}) \in [0,1]^{3}$ for $j \in \{1:D_{x^{te}}\}$, where $(p^j_{a},p^j_{n},p^j_{u})=:(g_{\theta^*}(s^{te}_j)_{\{a\}},g_{\theta^*}(s^{te}_j)_{\{n\}},g_{\theta^*}(s^{te}_j)_{\{u\}})$ and $D_{x^{te}}$ is the total number of sentences in the report $x^{te}$. Here, $p^j_{a}$, $p^j_{n}$, and $p^j_{u}$ represent the sentence-level (the $j$-th sentence's) probability for being abnormal, normal, and uncertain, respectively. Then, the final document-level abnormality probability $p_{a}$ is driven as the highest sentence-level probability (togetherwith its inverse value as normal probability $p_{n}$) as follows:
\begin{align} 
 p_{a} &\leftarrow  {\max}_{(j \in \{1:D_{x^{te}}\})} \, \big[ g_{\theta^*}(s^{te}_j)_{\{a\}} \big], \\\nonumber 
p_{n} &\leftarrow  1 - p_{a}.
\end{align}
This allows for an abnormal document classification if even one sentence is deemed abnormal.
\end{itemize}
 
\begin{table}[h]
  \centering     

  {\resizebox{0.6\linewidth}{!}{%
  \renewcommand{\arraystretch}{1.15}
  {\begin{tabular}{cccccc} 
  \toprule
  \textbf{Percentage(\%)} & \textbf{GT} & \textbf{Abnormal} & \textbf{Normal} & \textbf{Uncertain} & \begin{tabular}[c]{@{}c@{}}\textbf{Doc.}\\\textbf{Count}\end{tabular} \\ 
  \toprule
  \multirow{2}{*}{\begin{tabular}[c]{@{}c@{}}\\\vspace{0.8\baselineskip} Test Dataset\end{tabular}} & Abnor & \begin{tabular}[c]{@{}c@{}}\textbf{44.85}\\\textbf{± 22.1}\end{tabular} & \begin{tabular}[c]{@{}c@{}}27.17\\± 20.1\end{tabular} & \begin{tabular}[c]{@{}c@{}}27.97\\± 17.3\end{tabular} & 2394 \\
  & Normal & \begin{tabular}[c]{@{}c@{}}0.0\\± 0.0\end{tabular} & \begin{tabular}[c]{@{}c@{}}63.61\\± 21.1\end{tabular} & \begin{tabular}[c]{@{}c@{}}\textbf{36.39}\\\textbf{± 21.1}\end{tabular} & 438 \\ 
  \midrule
   \multirow{2}{*}{\begin{tabular}[c]{@{}c@{}}\\\vspace{0.8\baselineskip} D-KD Incorrect\end{tabular}} & Abnor & \begin{tabular}[c]{@{}c@{}}\textbf{29.14}\\\textbf{± 17.4}\end{tabular} & \begin{tabular}[c]{@{}c@{}}42.17\\± 21.1\end{tabular} & \begin{tabular}[c]{@{}c@{}}28.69\\± 18.0\end{tabular} & 340 \\
   & Normal & \begin{tabular}[c]{@{}c@{}}0.0\\± 0.0\end{tabular} & \begin{tabular}[c]{@{}c@{}}61.86\\± 26.3\end{tabular} & \begin{tabular}[c]{@{}c@{}}\textbf{38.14}\\\textbf{± 26.3}\end{tabular} & 70 \\ 
    \midrule
    \multirow{2}{*}{\begin{tabular}[c]{@{}c@{}}\\\vspace{0.8\baselineskip} S-KD Incorrect\end{tabular}} & Abnor & \begin{tabular}[c]{@{}c@{}}\textbf{21.28}\\\textbf{± 11.5}\end{tabular} & \begin{tabular}[c]{@{}c@{}}51.10\\± 17.3\end{tabular} & \begin{tabular}[c]{@{}c@{}}27.62\\± 18.3\end{tabular} & 114 \\
    & Normal & \begin{tabular}[c]{@{}c@{}}0.0\\± 0.0\end{tabular} & \begin{tabular}[c]{@{}c@{}}58.79\\± 27.4\end{tabular} & \begin{tabular}[c]{@{}c@{}}\textbf{41.21}\\\textbf{± 27.4}\end{tabular} & 26 \\ 
    \midrule
    \multirow{2}{*}{\begin{tabular}[c]{@{}c@{}} D-KD Incorrect\\$\cap$\\ S-KD Correct\end{tabular}} & Abnor & \begin{tabular}[c]{@{}c@{}}31.46\\± 18.0\end{tabular} & \begin{tabular}[c]{@{}c@{}}40.01\\± 20.9\end{tabular} & \begin{tabular}[c]{@{}c@{}}28.53\\± 17.8\end{tabular} & 282 \\
    & Normal & \begin{tabular}[c]{@{}c@{}}0.0\\± 0.0\end{tabular} & \begin{tabular}[c]{@{}c@{}}63.69\\± 24.5\end{tabular} & \begin{tabular}[c]{@{}c@{}}36.31\\± 24.5\end{tabular} & 60 \\
    \toprule
    \end{tabular}
    }}}
  \caption{Distribution comparison in medical reports: Analyzing abnormal, normal, and uncertain sentences between D-KD and S-KD on RadBERT-Roberta}
  \label{tab:distri_sentence}
\end{table}
 
\subsection{Objective Function for KD Training}\label{sec:kd_loss} Note the KD objective function $\mathcal{L}_{\theta}$ used in Eqs. \eqref{loss_doc} and \eqref{loss_sent}, is defined as
\begin{align}\label{total_loss}
\mathcal{L}_{\theta}(g_{\theta}(x),y):= \mathcal{L}^{cross}_{\theta}(g_{\theta}(x),y) + \lambda \cdot L^{cont}_{\theta}(g_{\theta}(x),y),
\end{align}
by adding the supervised contrastive loss $L^{cont}$ \citep{khosla2020supervised} to the cross-entropy loss $\mathcal{L}^{cross}$, where
\begin{align}\nonumber
\mathcal{L}^{cross}_{\theta}(g_{\theta}(x),y) &:= -\sum_{k} p_y[k] \cdot \log\Big(g_{\theta}(x)[k]\Big), \\\nonumber 
L^{cont}_{\theta}(g_{\theta}(x),y) &:= - \log \mathbb{E}_{(v \in B_y)} \Bigg[ \frac{e^{(\text{sim}(z, z_{v})/\tau)}}{\sum\limits_{k \in B} e^{(\text{sim}(z, z_k)/\tau)}}\Bigg].
\end{align} 
Here, $p_y$ is the one-hot vector representation of $y$, $z:=z_{\theta}(x)$ and $z_v:=z_{\theta}(x_v)$ represent the latent feature vectors of $g_{\theta}$ for the target input $x$ and another $x_v$ (where $y_v$ is its label), $B$ is the batch set of training data (and $B_y:=\{v \in B|y_v=y\}$ is its subset whose labels are our target label $y$), and $\text{sim}(\cdot)$ is the similarity metric. The addition of the contrastive loss $L^{cont}$ aims to minimize distances in the latent space $z_{\theta}(\cdot)$ within the same class and maximize those between different classes, thereby enhancing KD performance by strengthening the balance of each class (refer to Secs. \ref{sec:ablation_contrastive} and \ref{sec:cause_contrastive}).
 
\begin{figure}[t!]
      \centering
      {%
      \subfigure[Abnormal sentence distribution for abnormal documents]{\label{fig:distribution_abnormal}%
        \includegraphics[width=0.8\linewidth]{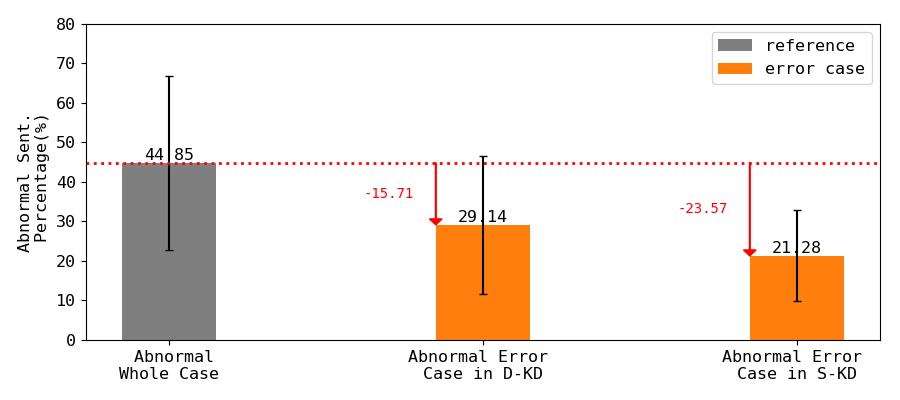}} 
      \subfigure[Uncertain sentence distribution for normal documents]{\label{fig:distribution_normal}%
        \includegraphics[width=0.8\linewidth]{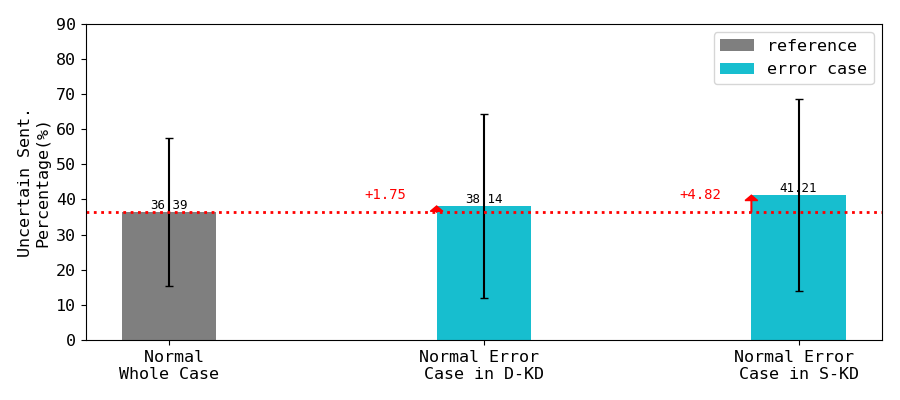}}
      }
      \captionof{figure}{Comparison of AD performance between S-KD and D-KD: S-KD demonstrates superior detection in abnormal (or normal) documents with fewer abnormal (or uncertain) sentences, outperforming D-KD in identifying challenging AD cases}
      \label{fig:distribution}
\end{figure}
 
\begin{figure*}[t!] 
\centering
{\includegraphics[width=0.95\linewidth]{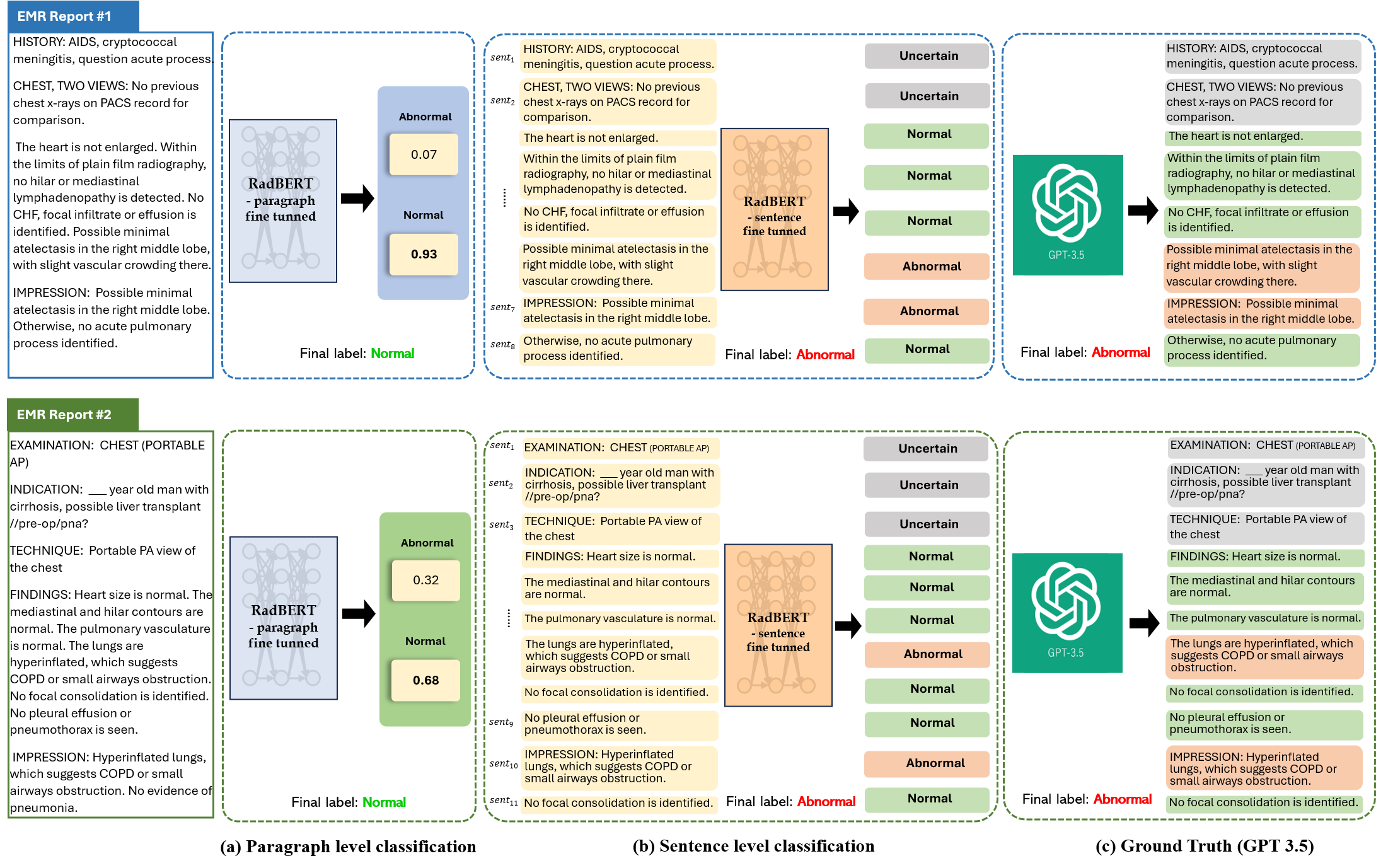}}
  \caption{Sample cases: Document-level vs Sentence-level KD - Demonstrating instances where document-level KD fails and sentence-level KD succeeds in accurately predicting abnormal and normal medical reports}
  \label{fig:KD_reasoning}
\end{figure*}
 
\section{Results}
\label{sec:res}
In this section, we present the experimental results of our study. Specifically, we explain the superiority of sentence-level knowledge distillation (S-KD) over the traditional document-level KD (D-KD) (Sec. \ref{sec:comparision_d&s}), along with the underlying reasons (Sec. \ref{sec:casue_s-kd}). Additionally, we discuss the advantages of integrating a contrastive setup into KD training (Sec. \ref{sec:casue_s-kd}) and the rationale behind this enhancement (Sec. \ref{sec:cause_contrastive}). The setup details are given in Sec. \ref{sec:setup}.
 
\subsection{Setup}
\label{sec:setup} 
We employed the MIMIC-CXR dataset \citep{johnson2019mimic} for both our training and test datasets. For the training dataset, we used all of the p10 documents. For the test dataset, we only used the initial subset of the p11 documents. We employed GPT-3.5 to assign labels (normal $n$, abnormal $a$, and uncertain $u$) to each sentence in the datasets. Documents were labeled based on the presence of abnormal sentences; if any abnormal sentences were detected within a document, the document was classified as abnormal, otherwise, it was labeled as normal.

Using our high-confidence label selection method given in Appendix, we extract documents and sentences with high confidence from the target dataset; 11,158 training documents (1,698 normal and 9,860 abnormal) and 2,832 testing documents (2,394 abnormal and 438 normal). These documents consist of 172,105 training sentences (51,568 normal, 64,715 abnormal, and 55,822 uncertain) and 40,779 testing sentences (12,105 normal, 15,655 abnormal, and 13,019 uncertain). Baseline D-KD training uses the document-level dataset, whereas our S-KD training employs the associated sentence-level dataset for a fair comparison. Other details are in Appendix.

\subsection{Performance Comparison between Proposed and Baseline KD Approach}
\label{sec:comparision_d&s} 
In this section, we present a comparative analysis of the test performances of two knowledge distillation (KD) methods as outlined in Sec. \ref{sec:method}: document-level KD (D-KD) and sentence-level KD (S-KD). These methodologies were applied to six medically specialized BERT (Bidirectional Encoder Representations from Transformers)-based backbone models as for KD student models. The models assessed under KD training are RadBERT-Roberta \citep{yan2022radbert}, BioMed-Roberta \citep{gururangan2020don't}, BioBERT \citep{lee2019biobert}, ClinicalBERT \citep{alsentzer2019publicly}, BiomedBERT \citep{pubmedbert}, and BlueBERT \citep{peng2019transfer}.

Our results, depicted in Table \ref{tab:comp_d_s}, reveal the statistical significance of S-KD's performance superiority over the baseline D-KD across all evaluated models. This was corroborated by the Wilcoxon rank-sum test, indicating p-values less than 0.05 for various evaluation metrics. S-KD consistently showed improved performance in terms of accuracy, specificity, sensitivity, and AUC, with average enhancements of 1.047 times, 1.053 times, 1.053 times, and 1.051 times, respectively. It is noteworthy that metrics other than AUC were measured at the optimal threshold on the AUC curve.

Particularly noteworthy is the performance of the recently introduced RadBERT model. In our evaluations, the anomaly detection accuracy of the S-KD method on RadBERT was recorded at 95.06\%. This represents a substantial reduction in error rate, approximately threefold, compared to the D-KD method, which achieved an accuracy of 85.52\%. These results underscore the superiority and efficacy of the S-KD technique in this context.

\begin{table}[t]
\centering

  {\resizebox{0.6\linewidth}{!}{%
  {\begin{tabular}{ccccc}
  \toprule
  \textbf{Model}                                                                              & \textbf{Accuracy}                                       & \textbf{Specificity}                                    & \textbf{Sensitivity}                                    & \textbf{AUC}                                              \\ 
    \toprule
    \begin{tabular}[c]{@{}c@{}}RadBERT-Roberta\\-4m-document\\\textbf{baseline}\end{tabular}    & 85.17                                                   & 0.832                                                   & \textbf{0.852}                                                   & 0.846                                                     \\
    \begin{tabular}[c]{@{}c@{}}RadBERT-Roberta\\-4m-document\\\textbf{contrastive}\end{tabular} & \begin{tabular}[c]{@{}c@{}}\textbf{85.52}\\(+ 0.35)\end{tabular} & \begin{tabular}[c]{@{}c@{}}\textbf{0.858}\\(+0.026)\end{tabular} & \begin{tabular}[c]{@{}c@{}}0.840\\(-0.012)\end{tabular} & \begin{tabular}[c]{@{}c@{}}\textbf{0.901}\\(+0.055)\end{tabular}   \\ 
    \midrule
    \begin{tabular}[c]{@{}c@{}}RadBERT-Roberta\\-4m-sentence\\\textbf{baseline}\end{tabular}    & 91.53                                                   & 0.910                                                   & 0.936                                                   & 0.962                                                     \\
    \begin{tabular}[c]{@{}c@{}}RadBERT-Roberta\\-4m-sentence\\\textbf{contrastive}\end{tabular} & \begin{tabular}[c]{@{}c@{}}\textbf{95.06}\\(+3.53)\end{tabular}  & \begin{tabular}[c]{@{}c@{}}\textbf{0.941}\\(+0.031)\end{tabular} & \begin{tabular}[c]{@{}c@{}}\textbf{0.952}\\(+0.016)\end{tabular} & \begin{tabular}[c]{@{}c@{}}\textbf{0.977}\\(+0.015)\end{tabular}   \\ 
    \midrule
    \begin{tabular}[c]{@{}c@{}}BioMed-Roberta-\\document\\\textbf{baseline}\end{tabular}        & 82.17                                                   & 0.814                                                   & \textbf{0.861}                                                   & \textbf{0.889}                                                     \\
    \begin{tabular}[c]{@{}c@{}}BioMed-Roberta-\\document\\\textbf{contrastive}\end{tabular}     & \begin{tabular}[c]{@{}c@{}}\textbf{86.12}\\(+3.95)\end{tabular}  & \begin{tabular}[c]{@{}c@{}}\textbf{0.869}\\(+0.055)\end{tabular} & \begin{tabular}[c]{@{}c@{}}0.820\\(-0.041)\end{tabular} & \begin{tabular}[c]{@{}c@{}}0.877\\(-0.012)\end{tabular}   \\ 
    \midrule
    \begin{tabular}[c]{@{}c@{}}BioMed-Roberta-\\sentence\\\textbf{baseline}\end{tabular}        & 94.42                                                   & 0.910                                                   & \textbf{0.961}                                                   & 0.976                                                     \\
    \begin{tabular}[c]{@{}c@{}}BioMed-Roberta-\\sentence\\\textbf{contrastive}\end{tabular}     & \begin{tabular}[c]{@{}c@{}}\textbf{94.60}\\(+0.18)\end{tabular}  & \begin{tabular}[c]{@{}c@{}}\textbf{0.947}\\(+0.037)\end{tabular} & \begin{tabular}[c]{@{}c@{}}0.943\\(-0.018)\end{tabular} & \begin{tabular}[c]{@{}c@{}}\textbf{0.979}\\(+ 0.003)\end{tabular}  \\ 
    \midrule
    \begin{tabular}[c]{@{}c@{}} p-value\end{tabular}                 & 0.021                                                  & 0.021                                                   & 0.282                                                   & 0.282                                                     \\
    \toprule 
    \end{tabular}
    }}}
  \caption{Anomaly detection performance comparison between with (ours) and without contrastive learning}
  \label{tab:comp_contra}
    \end{table}

\subsection{Analysis of Potential Cause for S-KD Advancement}
\label{sec:casue_s-kd}
 
Sec. \ref{sec:comparision_d&s} demonstrates that sentence-level knowledge distillation (S-KD) surpasses document-level knowledge distillation (D-KD) in performance. This section delves into the underlying reasons for this improved performance and presents the following key findings: S-KD exhibits superior capabilities over D-KD in two critical aspects. Firstly, S-KD more accurately identifies documents as abnormal when they contain only a low presence of abnormal sentences, which is a more challenging scenario for anomaly detection. Secondly, S-KD effectively corrects misclassifications where documents are incorrectly identified as abnormal due to a high number of uncertain sentences, despite being normal as truth.

To arrive at these conclusions, we analyzed the distribution of normal, abnormal, and uncertain sentences in medical reports. This comprehensive analysis encompassed the document cases of the entire test dataset, incorrect classifications by D-KD, incorrect classifications by S-KD, and cases where D-KD failed but S-KD succeeded, as illustrated in Fig. \ref{fig:distribution}.
 
\begin{figure}[t] 
\centering
  {%
   \subfigure[D-KD Latent Representation]{
    \includegraphics[width=0.6\linewidth]{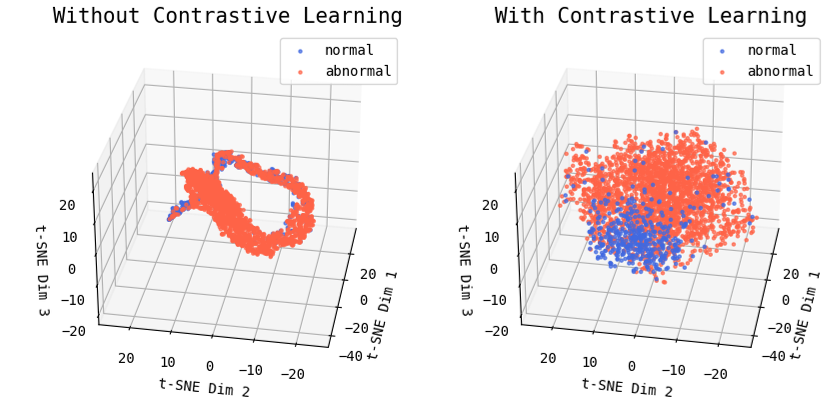}} 
  \subfigure[S-KD Latent Representation]{
    \includegraphics[width=0.6\linewidth]{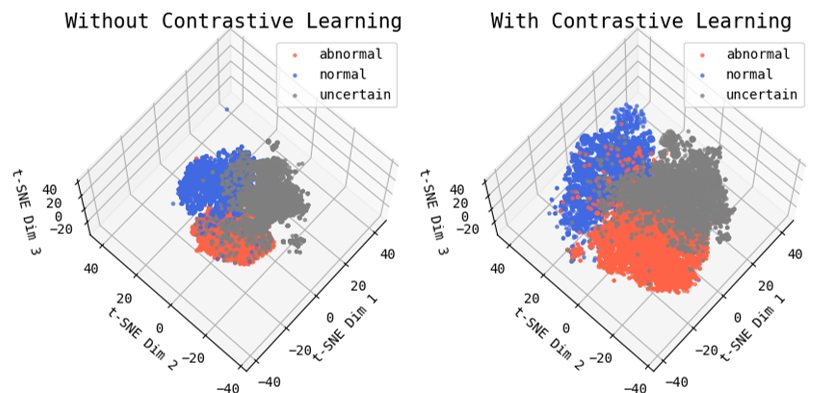}}
  }
  \caption{Comparison of latent vector distribution for each class depending on whether the contrastive setup is used (ours) or not}
  \label{fig:t-sne}
\end{figure}

In the document group where D-KD incorrectly classified cases from abnormal truth (Fig. \ref{fig:distribution_abnormal}), the distribution of abnormal sentences was approximately 29.14\%, compared to an average of 44.85\% in all abnormal truth documents. This indicates D-KD's struggle with sparsely abnormal sentences. In contrast, S-KD's incorrect cases showed a 21.28\% distribution of abnormal sentences, indicating higher accuracy with even fewer abnormal sentences. This suggests S-KD's enhanced capability in detecting sparser abnormal sentences by approximately 27\%. Notably, in cases where D-KD was incorrect but S-KD was correct, the document count was 282, signifying that S-KD correctly addressed 82.94\% of these D-KD incorrect cases (282 out of 340).

Regarding normal truth document cases (Fig. \ref{fig:distribution_normal}), documents incorrectly classified by D-KD had a 38.14\% distribution of uncertain sentences, higher than the test dataset's average of 36.39\%. This suggests that D-KD tends to be incorrect for documents where the number of uncertain sentences is large. In contrast, documents incorrectly classified by S-KD showed a 41.21\% distribution of uncertain sentences, effectively managing correct classifications with about 8.05\% more uncertain sentences. This emphasizes the impact of our consideration in training for sentence-level uncertainty as an additional/explicit label in S-KD. In particular, in the intersection of D-KD's incorrect and S-KD's correct cases, the document count was 60, indicating that S-KD correctly resolved 85.71\% of these D-KD incorrect document cases (60 out of 70).

Fig. \ref{fig:KD_reasoning} presents examples that illustrate the two primary reasons for the enhanced performance of S-KD. In the first document, a relatively small proportion, approximately 18.18\% (i.e., 2 out of 11), consists of abnormal sentences. Under the D-KD approach, accurately detecting these sparse abnormal sentences poses a challenge, leading to incorrect classifications. In contrast, the S-KD approach effectively identifies these sparse abnormal sentences, resulting in correct classification. In the second report, uncertain sentences constitute 40\% (i.e., 4 out of 10) of the document. Specifically, sentences 2 and 7 contribute to the ambiguity regarding abnormality within the D-KD framework, culminating in an erroneous classification. However, S-KD successfully navigates this ambiguity, accurately classifying the report even amidst a substantial presence of uncertain sentences.

\subsection{Ablation Study for Contrastive Loss in KD Learning}
\label{sec:ablation_contrastive}

\begin{table}[t]
\centering

  {\resizebox{0.6\linewidth}{!}{%
  \renewcommand{\arraystretch}{1.3}
  {\begin{tabular}{cccccc} 
   \toprule
   \multicolumn{2}{c}{\multirow{2}{*}{\begin{tabular}[c]{@{}c@{}}KD Method\\With vs Without Cont.\end{tabular}}} & \multicolumn{2}{c}{\textbf{D-KD}} & \multicolumn{2}{c}{\textbf{S-KD} (ours)} \\ 
   \multicolumn{2}{c}{} & \begin{tabular}[c]{@{}c@{}}Without \\Cont.\end{tabular} & \begin{tabular}[c]{@{}c@{}}With \\Cont.\end{tabular} & \begin{tabular}[c]{@{}c@{}}Without \\Cont.\end{tabular} & \begin{tabular}[c]{@{}c@{}}With \\Cont.\end{tabular} \\ 
   \midrule
   \multirow{3}{*}{\begin{tabular}[c]{@{}c@{}}Error Distance \\ (Int./Ext. Dist.)\end{tabular}} & Normal & 1.39~ & \textbf{0.55~} & 0.75 & \textbf{0.67} \\
   & Abnormal & 1.30~ & \textbf{0.70~} & 0.66 & \textbf{0.63} \\
   & Uncertain & — & — & \textbf{0.68} & 0.75 \\
   \toprule
\end{tabular}
}}}
  \caption{Comparison of error distance depending on whether contrastive setup is used (ours) or not}
  \label{tab:contra_distance}
\end{table}

In our study, we formulated a KD training objective that incorporates a contrastive learning objective, $\mathcal{L}^{cont}$, in addition to the conventional cross-entropy loss, $\mathcal{L}^{cross}$, as detailed in Eq. \eqref{total_loss}. Specifically, for our baseline model, which excludes contrastive learning, we set the parameter $\lambda$ to 0. Conversely, in our enhanced KD training scheme that includes contrastive learning, we assigned a value of 1 to $\lambda$ in Eq. \eqref{total_loss}.

Our ablation study, presented in Table \ref{tab:comp_contra}, examines the effect of adding the contrastive learning objective. This study was performed using the two highest-performing backbone models for AD as identified in Table \ref{tab:comp_d_s}: RadBERT-Roberta and BioMED-Roberta. The results, as depicted in Table \ref{tab:comp_contra}, demonstrate that incorporating the contrastive setting ($\lambda=1$) significantly improves accuracy and specificity in AD testing for all student models, compared to the baseline ($\lambda=0$) which does not include contrastive learning. However, the baseline setting exhibits a higher specificity, a discrepancy we attribute to the training data distribution. As discussed in Sec. \ref{sec:setup}, the training data contains a higher proportion of abnormal sentences and documents compared to normal instances, leading to a baseline bias towards abnormal labels. This results in elevated sensitivity but reduced specificity. The introduction of the contrastive setting helps to counteract this bias, enhancing specificity and thereby improving overall accuracy relative to the baseline. Consequently, we applied contrastive learning in all experiments except this ablation study.
 
\subsection{Analysis of Potential Cause for Contrastive Setting Advancement}
\label{sec:cause_contrastive}
In the previous section, our investigation centered on the enhanced capacity of our network to accurately identify the normal class, a minor category, through the application of contrastive learning. This approach resulted in a significant improvement in specificity, suggesting that contrastive learning contributes to more precise clustering of feature vectors within the same class. The current section aims to provide both qualitative and quantitative evidence to further substantiate the performance improvements attributable to the use of contrastive learning.

For visual validation, we extracted latent feature vectors from the point immediately preceding the linear layers in our trained network for each sample in the test dataset. These vectors were then visualized using t-SNE, as shown in Fig. \ref{fig:t-sne}. The resultant visualization indicates a more pronounced demarcation between class clusters when contrastive learning is employed. Specifically, in the D-KD scenario, we observed a distinct separation between the blue and red class clusters. In the S-KD context, there was a marked decrease in the instances of gray class samples overlapping with the red class region. These observations underscore the efficacy of contrastive learning in enhancing class discriminability in our network.

We defined $c$ as the centroid for the samples of each class. The intra distance for a sample was calculated as the $\ell_2$ distance from its class centroid $c$, while the extra distance was determined as the average distance from $c$ to the centroids of other classes. We then computed an error distance for each sample, defined as the ratio of intra to extra distance, to gauge the proximity of a sample to its true class, with lower values indicating a closer alignment. The outcomes of these error distance calculations are in Table \ref{tab:contra_distance}; the application of contrastive learning significantly reduces the mean error distance for test samples in both normal and abnormal classes in both D-KD and S-KD scenarios. This reduction in error distance markedly enhances the network's capability to detect even minor (i.e., normal) class, thereby improving the specificity and accuracy of the model's AD performance.
 
\begin{figure}[t] 
\centering
{\includegraphics[width=0.8\linewidth]{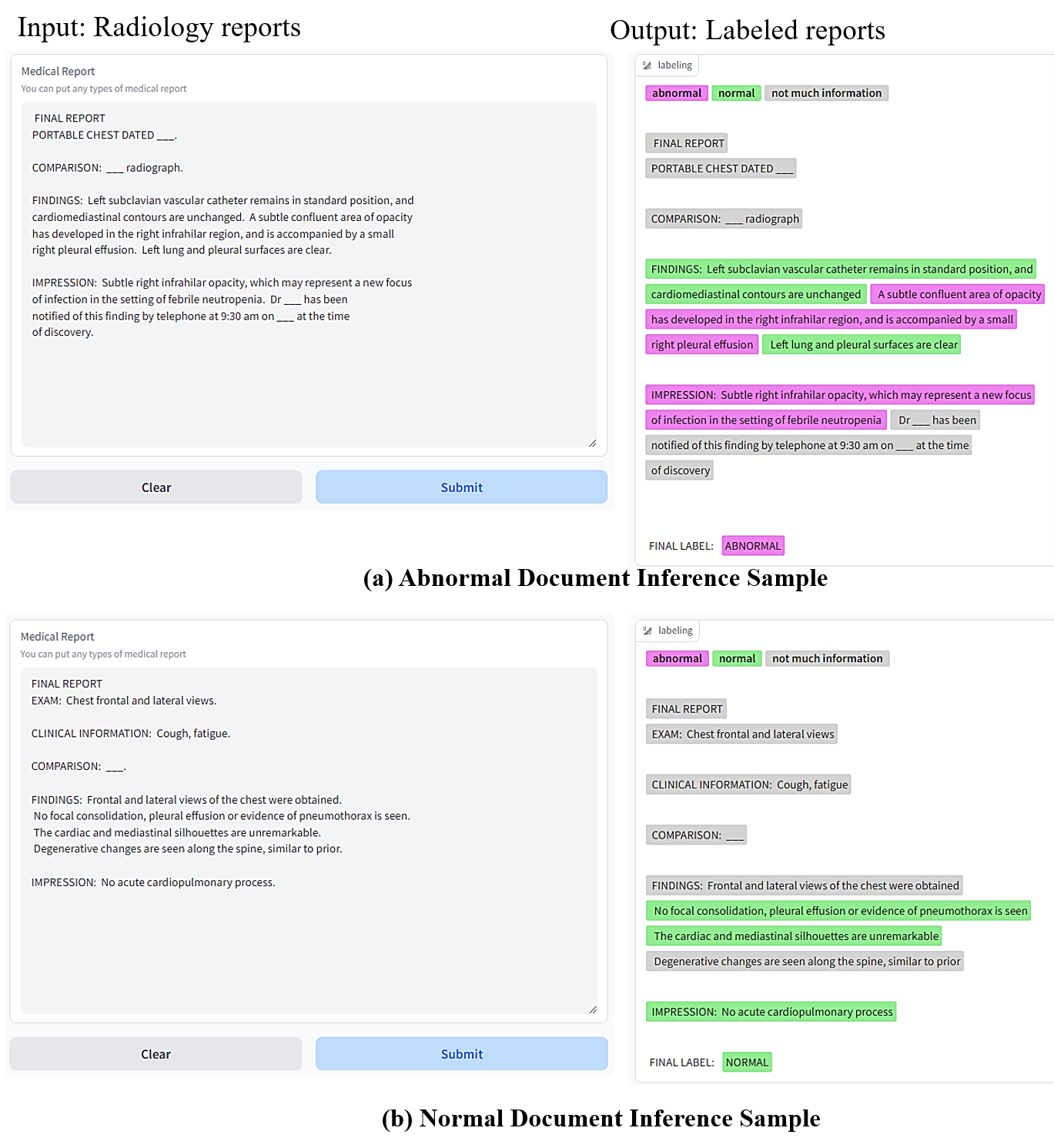}}
  \caption{Sample results of our model deployed on HuggingFace  - Demonstrating practical examples using color coding for enhanced interpretability}
  \label{fig:hugf}
\end{figure}

\section{Discussion}
\label{sec:dis}
In this section, we delineate the methodologies employed for extracting high-confidence labels from ChatGPT (Sec. \ref{sec:gpt_label}), detail the process of model deployment along with its potential utility (Sec. \ref{sec:model_deploy}), and discuss the limitations of our approach (Sec. \ref{sec:limitation}).

\subsection{Method to Extract High-Confidence Label from GPT 3.5} 
\label{sec:gpt_label}
 
To ascertain high-confidence labels from GPT-3.5, our approach involves an ensemble methodology. We execute three independent label extractions (normal, abnormal, uncertain) for each sentence from GPT-3.5. If these labels exhibit consistency across extractions, we accept them; otherwise, we reject them. Comprehensive details are presented in Appendix.

\subsection{Model Deployment and Implication}
\label{sec:model_deploy}
Our RadBERT-Roberta model, which was trained using sentence-level knowledge distillation (S-KD) with a contrastive setup, has been deployed and is accessible via \href{https://huggingface.co/spaces/junhyun01/Onpremise_LLM_Normal_Detection}{HuggingFace}.

Fig. \ref{fig:hugf} illustrates the model's functionality. When a clinician uploads a radiology report, the model highlights sentences indicative of normal and abnormal findings in green and purple, respectively, while sentences deemed uncertain are marked in gray. This feature enables radiologists or doctors to review reports more efficiently by focusing primarily on the text highlighted in green and purple, thereby potentially omitting the gray-marked uncertain content.

\subsection{Limitation}
\label{sec:limitation}  
First, our methodology was validated using the MIMIC-CXR dataset, a renowned public source, with distinct data separation for training. However, it lacks supplementary verification with varied public datasets. Future endeavors will expand this validation to encompass a broad spectrum of radiology reports.

Second, our study aims to address how to replicate the cloud-based model (i.e., ChatGPT-3.5) predictions in the non-cloud-based (i.e., secure) model. Therefore, GPT-3.5 was utilized for labeling radiologist reports without any human annotation load. However, our approach reveals a ground truth limitation: the absence of radiologist-confirmed ground truth in our process. Future work may focus on integrating radiologist-verified ground truth to enhance the accuracy of GPT-3.5's predictions. Nevertheless, our research demonstrates, for the first time, the potential of replicating ChatGPT within a secure model for radiologist report analysis (without any human manual annotation) and introduces advanced KD strategies tailored for this aim.

Lastly, our approach assumes the use of non-sensitive data (i.e., data without security issues) for training, as the training data still requires uploading to GPT-3.5. In practical applications, especially in hospital settings, this necessitates de-identification protocols to ensure data privacy. Despite this limitation, our method removes the need for human manual annotation (e.g., an indication of abnormal status per sentence), suggesting a promising direction for efficiently replicating cloud-based models like GPT-3.5 under in-hospital and secure environments. We also expect that the costs and efforts associated with de-identification are generally less than those required for manual annotation processes, thereby supporting the usefulness of our approach. Furthermore, our study is significant in that it utilizes only a limited portion of data as training material. This implies that all other unrestricted datasets can be used as evaluation data without security concerns, as test data do not need to be uploaded to GPT-3.5 but can be processed by our secure model. This capability presents a vital step forward in the practical application of AI in medical settings, offering a state-of-the-art (i.e., reproducing modern performance of cloud models like ChatGPT), efficient (i.e., without human annotation labor), and secure (i.e., implemented by non-cloud model) method for enhancing medical record processing.

\section{Conclusion}
\label{sec:con}
This paper presents a novel approach to replicating cloud model like ChatGPT as non-cloud one for secure usage in radiology report processing at hospitals, eliminating the need for human annotation. Our method involves a unique knowledge distillation process from ChatGPT, ensuring data remains on-site while maintaining comparable performance. The effectiveness of this approach is demonstrated through anomaly detection in radiology reports, highlighting our model's ability in sentence-level knowledge distillation and explicit management of  uncertainty. We also expect our approach's principle could extend to other report processing tasks such as question-answering tasks (e.g., for detection of individual disease), where our model's variant could adeptly identify and filter out low-confidence sentences in relation to the question. We expect that our research sets a precedent for developing secure and in-hospital LLM AI systems with minimal human supervision, potentially heralding a new era in healthcare technology applications.
  
\newpage
\appendix 

\section{Data Setup Details}\label{apd:data setup}

In our study using the MIMIC-CXR dataset \cite{johnson2019mimic}, we initially utilized a training set of 22,195 p10 documents and a test set of 5,217 p11 documents. Post our high-confidence label filtering (Sec. \ref{apd:gpt}), the training set, originally with 188,827 sentences, was reduced to 172,105, and the test set from 44,516 to 40,779 sentences. Only documents with all sentences classified as high-confidence were retained, resulting in 11,158 training (from 22,195) and 2,832 test (from 5,217) documents as the final dataset in our study.

The training set comprised 1,698 normal and 9,460 abnormal documents (i.e., total 11,158 documents), with 51,568 normal, 64,715 abnormal, and 55,822 uncertain sentence (i.e., total 172,105 sentence). The test set included 2,394 abnormal and 438 normal documents (i.e., total 2,832 documents), with 12,105 normal, 15,655 abnormal, and 13,019 uncertain sentences (i.e., total 40,779 sentence). This dataset enabled a robust comparison of our S-KD approach against the baseline D-KD.

\begin{figure*}[t] 
\centering
{\includegraphics[width=0.9\linewidth]{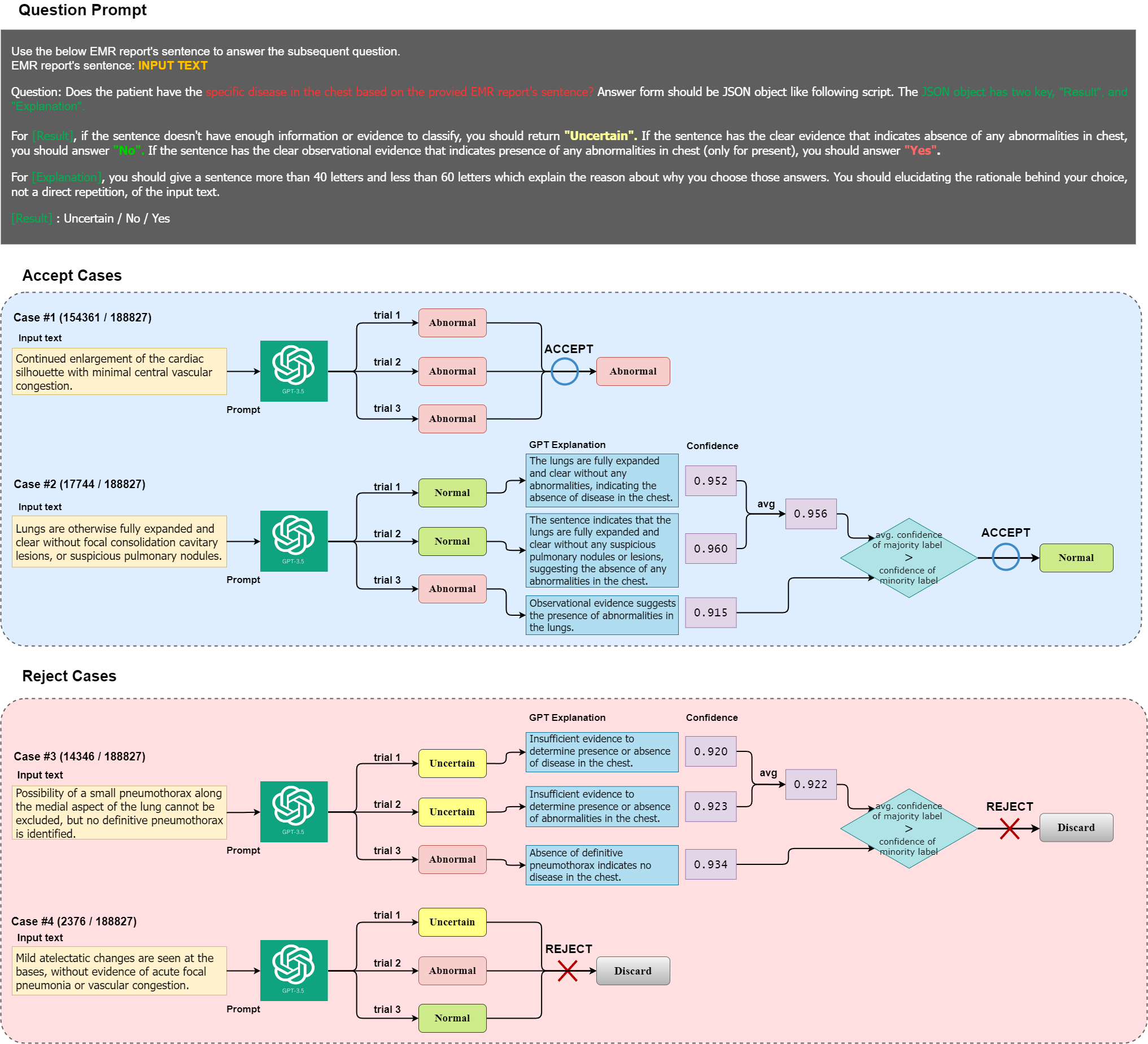}}
  \caption{Our approach employs two key techniques to derive high-confidence sentence labels from GPT-3.5: (1) the prompt engineering that obliges the network to elucidate the rationale behind its outputs, and (2) the prediction ensemble methodology to extract the result ensuring all models concur in providing identical reasoning.}
  \label{fig:gpt_method}
\end{figure*}

\section{High-Confidence Label Extraction from GPT-3.5}\label{apd:gpt}
Our research introduces a method for extracting high-confidence labels from GPT-3.5, as outlined in Fig. \ref{fig:gpt_method}. We obtain independent sentence labels from GPT-3.5 three times to ensure label consistency. Labels are considered high-confidence when all three extractions match. Discrepancies lead to label dismissal unless a majority (two out of three) consistency is observed. In such cases, we calculate the cosine similarity between the input text and the GPT  explanation for each label, followed by averaging the confidence scores of the consistent labels. A label is accepted if its average confidence score is higher than the score of the minority label. This method, crucial for extracting reliable labels from ChatGPT, serves as training data for knowledge distillation (KD), ensuring both label consistency and confidence based on GPT explanations.

\clearpage 

\begin{small}
\bibliographystyle{unsrtnat}
\bibliography{chil-sample.bib}

\begin{thebibliography}{24}
\providecommand{\natexlab}[1]{#1}
\providecommand{\url}[1]{\texttt{#1}}
\expandafter\ifx\csname urlstyle\endcsname\relax
  \providecommand{\doi}[1]{doi: #1}\else
  \providecommand{\doi}{doi: \begingroup \urlstyle{rm}\Url}\fi

\bibitem[Wu et~al.(2024)Wu, Koo, Blum, Black, Kao, Fei, Scalzo, and
  Kurtz]{wu2024benchmarking}
Sean Wu, Michael Koo, Lesley Blum, Andy Black, Liyo Kao, Zhe Fei, Fabien
  Scalzo, and Ira Kurtz.
\newblock {Benchmarking} open-source large language models, {GPT-4} and {Claude
  2} on multiple-choice questions in nephrology.
\newblock \emph{NEJM AI}, page AIdbp2300092, 2024.

\bibitem[Mirza et~al.(2024)Mirza, Tang, Connolly, Abdulrazeq, Lim, Roye,
  Priebe, Chandler, Libby, Groff, et~al.]{mirza2024using}
Fatima~N Mirza, Oliver~Y Tang, Ian~D Connolly, Hael~A Abdulrazeq, Rachel~K Lim,
  G~Dean Roye, Cedric Priebe, Cheryl Chandler, Tiffany~J Libby, Michael~W
  Groff, et~al.
\newblock {Using} {ChatGPT} to facilitate truly informed medical consent.
\newblock \emph{NEJM AI}, page AIcs2300145, 2024.

\bibitem[Lee et~al.(2023)Lee, Bubeck, and Petro]{lee2023benefits}
Peter Lee, Sebastien Bubeck, and Joseph Petro.
\newblock {Benefits}, limits, and risks of {GPT-4} as an {AI} chatbot for
  medicine.
\newblock \emph{New England Journal of Medicine}, 388\penalty0 (13):\penalty0
  1233--1239, 2023.

\bibitem[Senbekov et~al.(2020)Senbekov, Saliev, Bukeyeva, Almabayeva,
  Zhanaliyeva, Aitenova, Toishibekov, and Fakhradiyev]{senbekov2020}
Maksut Senbekov, Timur Saliev, Zhanar Bukeyeva, Aigul Almabayeva, Marina
  Zhanaliyeva, Nazym Aitenova, Yerzhan Toishibekov, and Ildar Fakhradiyev.
\newblock {The} recent progress and applications of digital technologies in
  healthcare: {A} review.
\newblock \emph{International journal of telemedicine and applications}, 2020,
  2020.

\bibitem[Gostin et~al.(2009)Gostin, Levit, Nass, et~al.]{gostin2009beyond}
Lawrence~O Gostin, Laura~A Levit, Sharyl~J Nass, et~al.
\newblock {Beyond} the {HIPAA} privacy rule: {E}nhancing privacy, improving
  health through research.
\newblock 2009.

\bibitem[Voigt and Von~dem Bussche(2017)]{voigt2017eu}
Paul Voigt and Axel Von~dem Bussche.
\newblock {T}he {EU} general data protection regulation {(GDPR)}.
\newblock \emph{A Practical Guide, 1st Ed., Cham: Springer International
  Publishing}, 10\penalty0 (3152676):\penalty0 10--5555, 2017.

\bibitem[Li et~al.(2023)Li, Moon, Iyer, Balthazar, Krupinski, Bercu, Newsome,
  Banerjee, Gichoya, and Trivedi]{li2023decoding}
Hanzhou Li, John~T Moon, Deepak Iyer, Patricia Balthazar, Elizabeth~A
  Krupinski, Zachary~L Bercu, Janice~M Newsome, Imon Banerjee, Judy~W Gichoya,
  and Hari~M Trivedi.
\newblock {Decoding} radiology reports: {P}otential application of {OpenAI}
  {ChatGPT} to enhance patient understanding of diagnostic reports.
\newblock \emph{Clinical Imaging}, 2023.

\bibitem[Liu et~al.(2023{\natexlab{a}})Liu, Zhong, Li, Zhang, Pan, Zhao, Dong,
  Cao, Liu, Shu, et~al.]{liu2023evaluating}
Zhengliang Liu, Tianyang Zhong, Yiwei Li, Yutong Zhang, Yi~Pan, Zihao Zhao,
  Peixin Dong, Chao Cao, Yuxiao Liu, Peng Shu, et~al.
\newblock {Evaluating} large language models for radiology natural language
  processing.
\newblock \emph{arXiv preprint:2307.13693}, 2023{\natexlab{a}}.

\bibitem[Ma et~al.(2023)Ma, Wu, Wang, Xu, Wei, Liu, Guo, Cai, Zhang, Zhang,
  et~al.]{ma2023impressiongpt}
Chong Ma, Zihao Wu, Jiaqi Wang, Shaochen Xu, Yaonai Wei, Zhengliang Liu, Lei
  Guo, Xiaoyan Cai, Shu Zhang, Tuo Zhang, et~al.
\newblock {ImpressionGPT:} {An} iterative optimizing framework for radiology
  report summarization with {chatGPT}.
\newblock \emph{arXiv preprint:2304.08448}, 2023.

\bibitem[Liu et~al.(2023{\natexlab{b}})Liu, Li, Shu, Zhong, Yang, Ju, Wu, Ma,
  Luo, Chen, Kim, Hu, Dai, Zhao, Zhu, Liu, Liu, Shen, Liu, Li, and
  Li]{Liu2023RadiologyLlama2BL}
Zheng Liu, Yiwei Li, Peng Shu, Aoxiao Zhong, Longtao Yang, Chao Ju, Zihao Wu,
  Chong-Yi Ma, Jie Luo, Cheng Chen, Sekeun Kim, Jiang Hu, Haixing Dai, Lin
  Zhao, Dajiang Zhu, Jun Liu, W.~Liu, Dinggang Shen, Tianming Liu, Quanzheng
  Li, and Xiang Li.
\newblock {Radiology-Llama2}: {B}est-in-class large language model for
  radiology.
\newblock \emph{ArXiv}, abs/2309.06419, 2023{\natexlab{b}}.

\bibitem[Liu et~al.(2023{\natexlab{c}})Liu, Zhong, Li, Yang, Ju, Wu, Ma, Shu,
  Chen, Kim, et~al.]{liu2023radiology}
Zhengliang Liu, Aoxiao Zhong, Yiwei Li, Longtao Yang, Chao Ju, Zihao Wu, Chong
  Ma, Peng Shu, Cheng Chen, Sekeun Kim, et~al.
\newblock {Radiology-GPT:} {A} large language model for radiology.
\newblock \emph{arXiv preprint:2306.08666}, 2023{\natexlab{c}}.

\bibitem[Zhong et~al.(2023)Zhong, Zhao, Zhang, Pan, Dong, Jiang, Kui, Shang,
  Yang, Wei, et~al.]{zhong2023chatradio}
Tianyang Zhong, Wei Zhao, Yutong Zhang, Yi~Pan, Peixin Dong, Zuowei Jiang,
  Xiaoyan Kui, Youlan Shang, Li~Yang, Yaonai Wei, et~al.
\newblock {Chatradio-valuer:} {A} chat large language model for generalizable
  radiology report generation based on multi-institution and multi-system data.
\newblock \emph{arXiv preprint:2310.05242}, 2023.

\bibitem[Van~Veen et~al.(2023)Van~Veen, Van~Uden, Attias, Pareek, Bluethgen,
  Polacin, Chiu, Delbrouck, Chaves, Langlotz, et~al.]{van2023radadapt}
Dave Van~Veen, Cara Van~Uden, Maayane Attias, Anuj Pareek, Christian Bluethgen,
  Malgorzata Polacin, Wah Chiu, Jean-Benoit Delbrouck, Juan Manuel~Zambrano
  Chaves, Curtis~P Langlotz, et~al.
\newblock {RadAdapt:} {R}adiology report summarization via lightweight domain
  adaptation of large language models.
\newblock \emph{arXiv preprint:2305.01146}, 2023.

\bibitem[Mukherjee et~al.(2023)Mukherjee, Hou, Lanfredi, and
  Summers]{mukherjee2023feasibility}
Pritam Mukherjee, Benjamin Hou, Ricardo~B Lanfredi, and Ronald~M Summers.
\newblock {Feasibility} of using the privacy-preserving large language model
  {Vicuna} for labeling radiology reports.
\newblock \emph{Radiology}, 309\penalty0 (1):\penalty0 e231147, 2023.

\bibitem[Bressem et~al.(2020)Bressem, Adams, Gaudin, Tr{\"o}ltzsch, Hamm,
  Makowski, Sch{\"u}le, Vahldiek, and Niehues]{bressem2020highly}
Keno~K Bressem, Lisa~C Adams, Robert~A Gaudin, Daniel Tr{\"o}ltzsch, Bernd
  Hamm, Marcus~R Makowski, Chan-Yong Sch{\"u}le, Janis~L Vahldiek, and Stefan~M
  Niehues.
\newblock {Highly} accurate classification of chest radiographic reports using
  a deep learning natural language model pre-trained on 3.8 million text
  reports.
\newblock \emph{Bioinformatics}, 36\penalty0 (21):\penalty0 5255--5261, 2020.

\bibitem[Yan et~al.(2022)Yan, McAuley, Lu, Du, Chang, Gentili, and
  Hsu]{yan2022radbert}
An~Yan, Julian McAuley, Xing Lu, Jiang Du, Eric~Y Chang, Amilcare Gentili, and
  Chun-Nan Hsu.
\newblock {RadBERT:} {Adapting} transformer-based language models to radiology.
\newblock \emph{Radiology: Artificial Intelligence}, 4\penalty0 (4):\penalty0
  e210258, 2022.

\bibitem[Gou et~al.(2021)Gou, Yu, Maybank, and Tao]{gou2021knowledge}
Jianping Gou, Baosheng Yu, Stephen~J Maybank, and Dacheng Tao.
\newblock {K}nowledge distillation: {A} survey.
\newblock \emph{International Journal of Computer Vision}, 129:\penalty0
  1789--1819, 2021.

\bibitem[Khosla et~al.(2020)Khosla, Teterwak, Wang, Sarna, Tian, Isola,
  Maschinot, Liu, and Krishnan]{khosla2020supervised}
Prannay Khosla, Piotr Teterwak, Chen Wang, Aaron Sarna, Yonglong Tian, Phillip
  Isola, Aaron Maschinot, Ce~Liu, and Dilip Krishnan.
\newblock {S}upervised contrastive learning.
\newblock \emph{34th Conference on Neural Information Processing Systems
  (NeurIPS2020), Vancouver, Canada}, 2020.

\bibitem[Johnson et~al.(2019)Johnson, Pollard, Berkowitz, Greenbaum, Lungren,
  Deng, Mark, and Horng]{johnson2019mimic}
Alistair E.~W. Johnson, Tom~J. Pollard, Seth~J. Berkowitz, Nathaniel~R.
  Greenbaum, Matthew~P. Lungren, Chih-ying Deng, Roger~G Mark, and Steven
  Horng.
\newblock {MIMIC-CXR,} a de-identified publicly available database of chest
  radiographs with free-text reports.
\newblock \emph{Scientific Data}, 6\penalty0 (317), 2019.

\bibitem[Gururangan et~al.(2020)Gururangan, Marasović, Swayamdipta, Lo,
  Beltagy, Downey, and Smith]{gururangan2020don't}
Suchin Gururangan, Ana Marasović, Swabha Swayamdipta, Kyle Lo, Iz~Beltagy,
  Doug Downey, and Noah~A. Smith.
\newblock {Don’t} stop pretraining: {Adapt} language models to domains and
  tasks.
\newblock \emph{Proceedings of the 58th Annual Meeting of the Association for
  Computational Linguistics}, page 8342–8360, 2020.

\bibitem[Lee et~al.(2020)Lee, Yoon, Kim, Kim, Kim, So, and
  Kang]{lee2019biobert}
Jinhyuk Lee, Wonjin Yoon, Sungdong Kim, Donghyeon Kim, Sunkyu Kim, Chan~Ho So,
  and Jaewoo Kang.
\newblock {BioBERT:} {A} pre-trained biomedical language representation model
  for biomedical text mining.
\newblock \emph{Bioinformatics}, 36:\penalty0 1234–1240, 2020.

\bibitem[Alsentzer et~al.(2019)Alsentzer, Murphy, Boag, Weng, Jindi, Naumann,
  and McDermott]{alsentzer2019publicly}
Emily Alsentzer, John Murphy, William Boag, Wei-Hung Weng, Di~Jindi, Tristan
  Naumann, and Matthew McDermott.
\newblock {Publicly} available clinical {BERT} embeddings.
\newblock \emph{Proceedings of the 2nd Clinical Natural Language Processing
  Workshop}, page 72–78, 2019.

\bibitem[Gu et~al.(2020)Gu, Tinn, Cheng, Lucas, Usuyama, Liu, Naumann, Gao, and
  Poon]{pubmedbert}
Yu~Gu, Robert Tinn, Hao Cheng, Michael Lucas, Naoto Usuyama, Xiaodong Liu,
  Tristan Naumann, Jianfeng Gao, and Hoifung Poon.
\newblock {Domain}-specific language model pretraining for biomedical natural
  language processing.
\newblock \emph{arXiv preprint:2007.15779}, 2020.

\bibitem[Peng et~al.(2019)Peng, Yan, and Lu]{peng2019transfer}
Yifan Peng, Shankai Yan, and Zhiyong Lu.
\newblock {Transfer} learning in biomedical natural language processing: {An}
  evaluation of {BERT} and {ELMo} on ten benchmarking datasets.
\newblock \emph{Proceedings of the 2019 Workshop on Biomedical Natural Language
  Processing (BioNLP 2019)}, pages 58--65, 2019.

\end{thebibliography}
\end{small} 
 

\end{document}